\documentclass[11pt,a4paper]{article}
\usepackage[hyperref]{emnlp2020}
\usepackage{times}
\usepackage{latexsym}

\usepackage{microtype}
\usepackage{url}
\usepackage{mathtools}
\usepackage{mathrsfs}
\usepackage{amsfonts}
\usepackage[linesnumbered,ruled,vlined]{algorithm2e}
\usepackage{multirow}
\usepackage{graphicx} 
\usepackage{float} 
\usepackage{subfigure} 
\usepackage{color}
\usepackage{enumerate}
\usepackage{booktabs}
\usepackage[normalem]{ulem}
\usepackage[toc,page,title]{appendix}
\usepackage{dcolumn}
\newcolumntype{d}[1]{D{,}{,}{#1}}

\aclfinalcopy 
\def\aclpaperid{2873} 

\usepackage{bm}

\newcommand\ourmodel{DiCGRL\xspace}

\title{Disentangle-based Continual Graph Representation Learning}

\author{Xiaoyu Kou\textsuperscript{1}\thanks{\ \ This work is done when Xiaoyu Kou was interning at
Pattern Recognition Center, WeChat AI, Tencent Inc, China},
Yankai Lin\textsuperscript{2},
Shaobo Liu\textsuperscript{2},
Peng Li\textsuperscript{2},
Jie Zhou\textsuperscript{2},
Yan Zhang\textsuperscript{1}\\
\textsuperscript{1}{Key Laboratory of Machine Perception (MOE),} \\
{Department of Machine Intelligence, Peking University, Beijing, China}\\
\textsuperscript{2}{Pattern Recognition Center, WeChat AI, Tencent Inc., China}\\
\texttt{\{kouxiaoyu,zhyzhy001\}@pku.edu.cn}\\
\texttt{\{yankailin,patrickpli,withtomzhou\}@tencent.com}
}
\date{}

\begin{document}
\maketitle
\begin{abstract}
Graph embedding (GE) methods embed nodes (and/or edges) in graph into a low-dimensional semantic space, and have shown its effectiveness in modeling multi-relational data. However, existing GE models are not practical in real-world applications since it overlooked the streaming nature of incoming data. To address this issue, we study the problem of \textit{continual graph representation learning} which aims to continually train a graph embedding model on new data to learn incessantly emerging multi-relational data while avoiding catastrophically forgetting old learned knowledge. Moreover, we propose a disentangle-based continual graph representation learning (\ourmodel) framework inspired by the human’s ability to learn procedural knowledge. The experimental results show that \ourmodel could effectively alleviate the catastrophic forgetting problem and outperform state-of-the-art continual learning models. 

\end{abstract}

\section{Introduction}
\label{sec_intro}
Multi-relational data represents relationships between entities in the world, which is usually denoted as a multi-relational graph with nodes and edges connecting them. It is widely used  in real-world NLP applications such as {\em knowledge graphs} (KGs) (e.g.,  Freebase~\cite{bollacker2008freebase} and DBpedia~\cite{lehmann2015dbpedia}) and  {\em information networks} (e.g., social network and citation network). Therefore, modeling multi-relational graph with graph embeddings~\citep{bordes2013translating,tang2015line,sun2019rotate,bruna2013spectral} has been attracting intensive attentions in both academia and industry. Graph  embedding (GE), aiming to embed nodes and/or edges in the graph into a low-dimensional semantic space to
enable neural models to effectively and efficiently utilize multi-relational data, 
has demonstrated remarkable effectiveness in various downstream NLP tasks such as question answering~\cite{bordes2014open} and dialogue system~\cite{moon2019opendialkg}.

\begin{figure}[!t]
	\centering
	\includegraphics[width=1.0\linewidth]{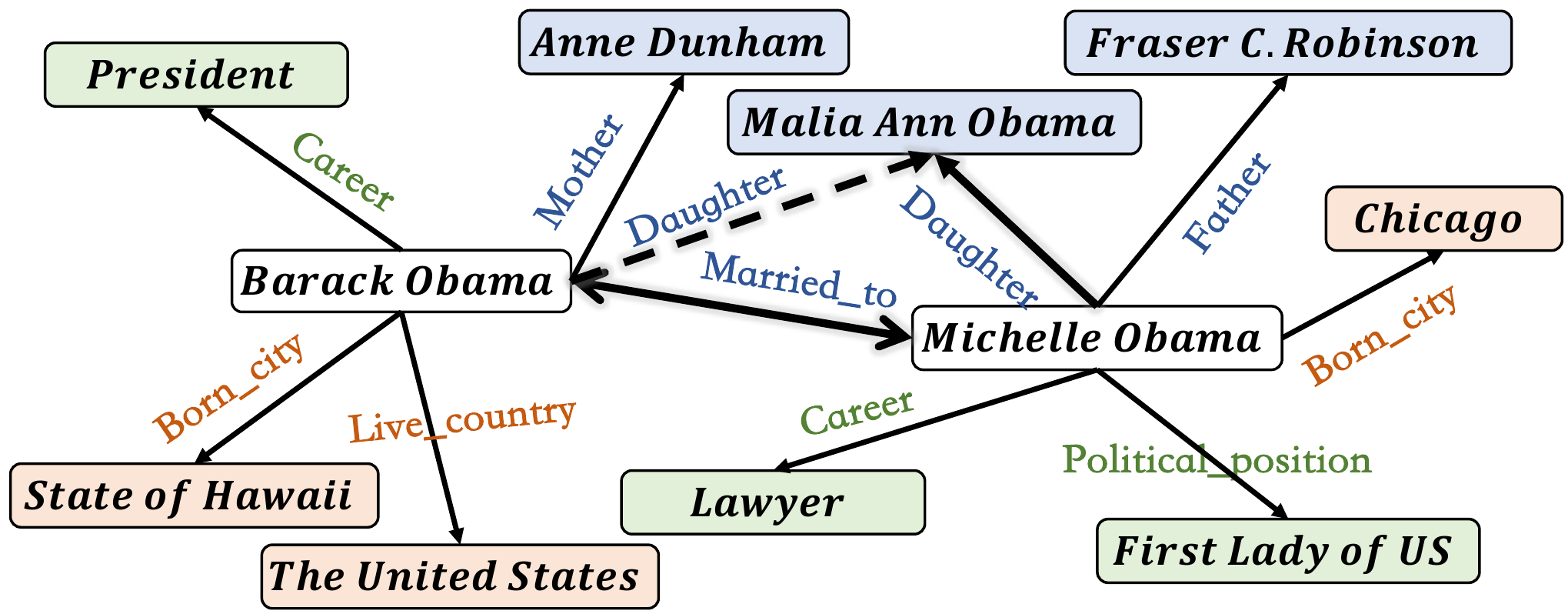}
	\caption{An example of edges that compressed into multiple components, which represented in different colors. The dotted line represents the inferred procedural knowledge from two \textbf{bold} edges.}
	\label{fig:intro}
	\vspace{-5mm}
\end{figure} 

Nevertheless, most existing graph embedding works overlook the streaming nature of the incoming data in real-world scenarios. 
In consequence, these models have to be retrained from scratch to reflect the data change, which is computationally expensive. To tackle this issue, we propose to study the problem of \textit{continual graph representation learning} (CGRL) in this work.

The goal of continual learning is to alleviate catastrophically forgetting old data while learning new data. There are two mainstream continual learning methods in NLP: (1) consolidation-based methods~\cite{kirkpatrick2017overcoming,zenke2017continual,liu2018rotate,ritter2018online} which consolidate the important model parameters of old data when learning new data; and (2) memory-based methods~\cite{lopez2017gradient,shin2017continual,chaudhry2019efficient} which remember a few old examples and learn them with new data jointly. Despite the promising results these methods have achieved on classification tasks, their effectiveness has not been validated on graph representation learning. 
Unlike the classification problem where instances are generally independent and can be operated individually,
nodes and edges in multi-relational data are correlated, making it sub-optimal to directly deploy existing continuous learning methods on the multi-relational data.

In cognitive psychology~\citep{solso2005cognitive}, procedural knowledge refers to a set of operational steps. Its smallest unit is production, where multiple productions can complete a series of cognitive activities. When learning new procedural knowledge, humans would update cognitive results by only updating a few related productions and leave the rest intact. Intuitively, such a process can be mimicked to learn constantly growing multi-relational data by regarding each new data as a new procedural knowledge. For example, as illustrated in Figure~\ref{fig:intro},  the 
relational triplets of \emph{Barack Obama} and \emph{Michelle Obama} are related to three concepts: ``family'', ``occupation'' and ``location''. When a new relational triplet (\emph{Michelle Obama}, \texttt{Daughter}, \emph{Malia Ann Obama}) appears, we only need to update the ``family''-related information in \emph{Barack Obama}. Consequently, we can further infer that the triplet (\emph{Barack Obama}, \texttt{Daughter}, \emph{Malia Ann Obama}) also holds. 

Inspired by procedural knowledge learning, we propose a disentangle-based continual graph representation learning framework \ourmodel in this work. Our proposed \ourmodel consists of two modules: (1)~\textbf{Disentangle module.} It decouples the relational triplets in the graph into multiple independent components according to their semantic aspects, and leverages two typical GE methods including Knowledge Graph Embedding (KGE) and Network Embedding (NE) to learn disentangled graph embeddings; (2) \textbf{Updating module.} When new relational triplets arrive, it selects the relevant old relational triplets and only updates the corresponding components of their graph embeddings. Compared with memory-based continual learning methods which save a fixed set of old data, \ourmodel could dynamically select important old data according to new data to fine-tune the model, which makes \ourmodel better model the complex multi-relational data stream. 

We conduct extensive experiments on both KGE and NE settings based on the real-world scenarios, and the experimental results show that \ourmodel effectively alleviates the catastrophic forgetting problem and significantly outperforms existing continual learning models while remaining efficient. 

\section{Related Work}

\subsection{Graph Embedding}
Graph embedding (GE) methods are critical techniques to obtain a good representation of multi-relational data.
There are mainly two categories of typical multi-relational data in the real-world, {\em knowledge graphs} (KGs) and  {\em information networks}. GE handles them via Knowledge Graph Embedding (KGE) and Network Embedding (NE) respectively, and our \ourmodel framework can adapt to the above two typical GE methods, which demonstrates the generalization ability of our model.

KGE is an active research area recently, which can be mainly divided into two categories to tackle link prediction task~\citep{ji2020survey}. One line of work is reconstruction-based models, which reconstruct the head/tail entity's embedding of a triplet using the relation and tail/head embeddings, such as TransE~\citep{bordes2013translating}, RotatE~\citep{sun2019rotate},  and ConvE~\citep{dettmers2018convolutional}.
Another line of work is bilinear-based models, which consider link prediction as a semantic matching problem. 
They take head, tail and relation's embeddings as inputs, and measure a semantic matching score for each triplet using bi-linear transformation (e.g., DistMult~\cite{yang2014embedding}, ComplEx~\cite{trouillon2016complex}, ConvKB~\cite{nguyen2017novel}). Besides KGE, NE is also widely explored in both academia and industry. Early works~\citep{10.1145/2623330.2623732,10.5555/3061053.3061163,10.1145/2736277.2741093} focus on learning static node embeddings on information graphs. More recently, graph neural networks~\cite{bruna2013spectral, henaff2015deep,velivckovic2017graph} have been attracting considerable attention and achieved remarkable success in learning network embeddings. 
However, most existing GE models assume the training data is static, i.e., do not change over time, which makes them impractical in real-world applications. 

\subsection{Continual Learning}
Continual learning, also known as life-long learning, helps alleviate catastrophic forgetting and enables incremental training for stream data. Methods for continual learning in natural language processing (NLP) field can mainly be divided into two categories: (1)~consolidation-based methods~\cite{kirkpatrick2017overcoming,zenke2017continual}, which slow down parameter updating to preserve old knowledge, and (2)
~memory-based methods~\cite{lopez2017gradient,shin2017continual,chaudhry2019efficient, wang2019sentence}, which retain examples from old data for re-play upon learning the new data. Although continual learning has been widely studied in NLP~\citep{sun2019lamal} and computer vision~\cite{kirkpatrick2017overcoming}, its exploration on graph embedding is relatively rare. \citet{sankar2018dynamic} seek to train graph embedding on constantly evolving data. However, it assumes the timestamp information is known beforehand, which hinders its application to other tasks. \citet{song2018enriching} extend the idea of regulation-based methods to continually learn graph embeddings which straightforwardly limits parameter updating on only the embedding layer. It is therefore hard to generalize to more complex multi-relational data. Our proposed \ourmodel model is distinct from previous works in two aspects: (1)~Our method does not require pre-annotated timestamps, which make it more feasible in various types of multi-relational data; (2)~Inspired by procedural knowledge learning, we exploit disentanglement to conduct continual learning and achieve promising results.

\section{Methodology}

\subsection{Task Formulation and Overall Framework}

We represent multi-relational data as a  multi-relational graph $G = (V, E)$, where $V$, $E$ denote the node set and the edge set within a graph $G$, and $G$ can be formalized as a set of relational triplets $\{(u, r, v)\}\subseteq V \times E \times V$. Given a relational triplet $(u, r, v) \in G$, we denote the embeddings of them as $\bm{u}$, $\bm{v}\in \mathbb{R}^{d}$ and $\bm{r} \in \mathbb{R}^{l}$, where $d$ and $l$ indicate the vector dimension. 

Continual graph representation learning trains graph embedding (GE) models on constantly growing multi-relational data, where the $i$-th part of multi-relational data has its own training set $\mathcal{T}_{i}$, validation set $\mathcal{V}_{i}$, and query set $\mathcal{Q}_{i}$. The $i$-th training set is defined as a set of relational triplets, i.e., $\mathcal{T}_{i} = \{(u_1^{\mathcal{T}_{i}},r_1^{\mathcal{T}_{i}},v_1^{\mathcal{T}_{i}}),\ldots,(u_N^{\mathcal{T}_{i}},r_N^{\mathcal{T}_{i}}, v_N^{\mathcal{T}_{i}})\}$, where $N$ is the instance number of $\mathcal{T}_{i}$. The $i$-th validation and query sets are defined similarly. 
As new relational triplets emerges, continual graph representation learning requires GE models to achieve good results on all previous query sets. 
Therefore, after training on the $i$-th training set $\mathcal{T}_{i}$, GE models will be evaluated on $\tilde{\mathcal{Q}}_{i}=\bigcup_{j=1}^{i}\mathcal{Q}_{j}$ to measure whether they could well model both new and old multi-relational data. The evaluation protocol indicates that it will be more and more difficult for the model to achieve high performance as the emerging of new relational triplets.

In general, our model continually learns on the streaming data. Whenever there comes a new part of multi-relational data, \ourmodel will learn the new graph embeddings and meanwhile prevent catastrophically forgetting old learned knowledge through two procedures: (1)  \textbf{Disentangle module}. It decouples the relational triplets in the graph into multiple components according to their semantic aspects, and learns disentangled graph embeddings that divide node embeddings into multiple independent components where each component describes a semantic aspect of node;  (2) \textbf{Updating module}. When new relational triplets arrive, \ourmodel first activates the old relational triplets from previous graphs which have relevant semantic aspects with the new ones, and only updates the corresponding components of their graph embeddings.

\subsection{Disentangle Module}
\begin{figure}[!t]
	\centering
	\includegraphics[width=1.0\linewidth]{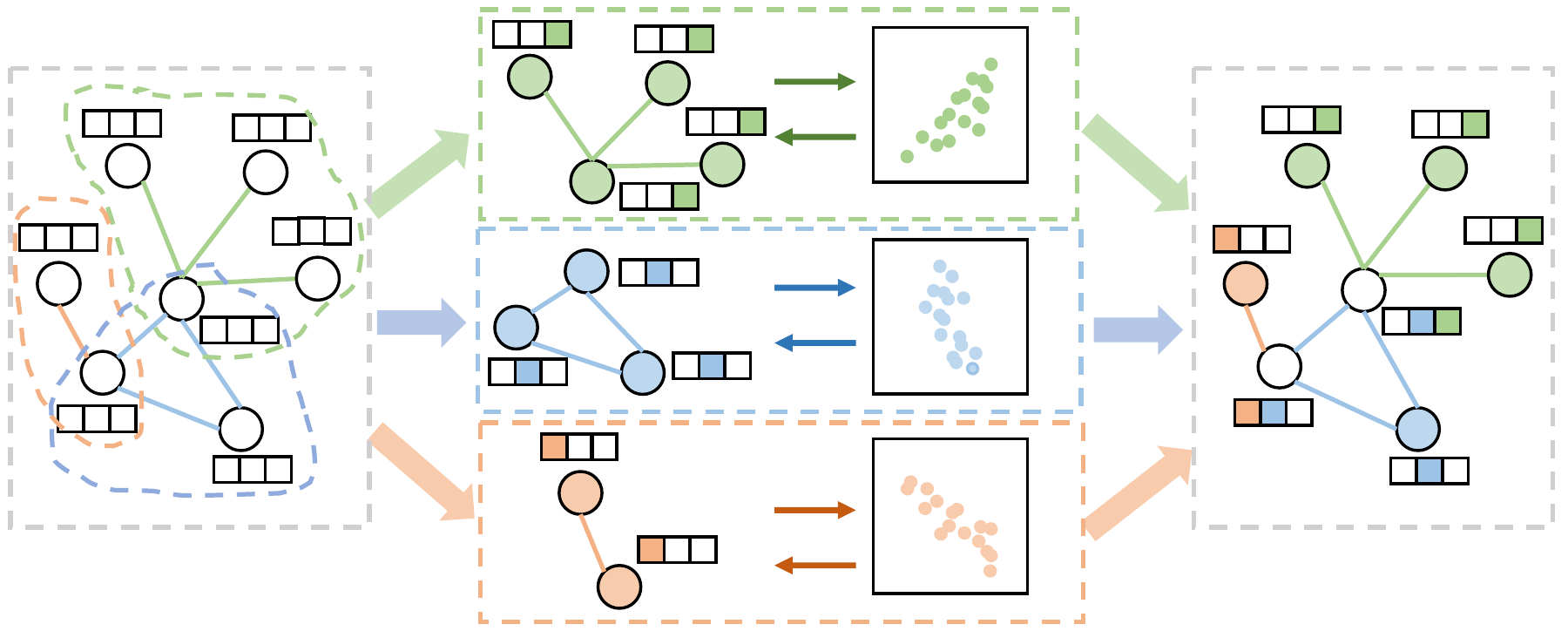}
	\caption{The disentangle module of our model. Different colors indicate different semantic aspects of nodes.}
	\label{fig:model_disen}
	\vspace{-3mm}
\end{figure}

When the $i$-th training set $\mathcal{T}_i$ becomes available, \ourmodel needs to update the graph embeddings according to these new relational triplets. 
To this end, for each node $u\in V$, we want to learn a disentangled node embedding $\bm{u}$, which is composed of $K$ independent components, i.e., $\bm{u} = [\bm{u}^1, \bm{u}^2,...,\bm{u}^k,...,\bm{u}^K]$, where $(0 \leq k \leq K)$ and $\bm{u}^k\in \mathbb{R}^{d}$. The component $\bm{u}^k$ is used to represent the $k$-th semantic aspect of node $u$.
As shown in Figure~\ref{fig:model_disen}, the key challenge of the disentangle module is how to decouple the relational triplets into multiple components according to their semantic aspects, and learn the disentangled graph embeddings in different components independently.

Formally, given a relational triplet $(u, r, v) \in \mathcal{T}_i$, we aim to extract the most related semantic components of $u$ and $v$ with respective to the relation $r$. Specifically, we model this process with an attention mechanism, where $(u, r, v)$ is associated with $K$ attention weight $(\alpha^1_r, \alpha^2_r, \ldots, \alpha^K_r)$, which respectively represent the probability being assigned to the $k$-th semantic component.  After that, we select the top-$n$ related components of $\bm{u}$ and $\bm{v}$ with the highest attention weight.
Then we leverage exiting GE methods to extract different features in the selected top-$n$ related components, and we denote the feature extraction operation as $\bm{f}$. Here, $\bm{f}$ could be any graph embedding operation that can incorporate the features of node $u$ and $v$ in the selected top-$n$ related components. In this work, we adapt our \ourmodel model in two typical graph embeddings including: 

\vspace{-0.5em}
\paragraph{Knowledge Graph Embeddings (KGEs)}

Intuitively, the most related semantic components of a relational triplet in KG are related to their relation $r$. Therefore, we can directly set $K$ attention values for each explicit relation $r$, and the $k$-th attention value $a^{k}_r\ (0 \leq k \leq K)$ is a trainable parameter which indicates how related this edge is to the $k$-th component. The normalized attention weight $\alpha^{k}_r$ is computed as: 
\begin{equation}
\label{eq:kg-att}
    \alpha^{k}_r = \frac{\exp(a^{k}_r)}{\sum_{j=1}^{K}\exp(a^{j}_r)}.
\end{equation}

As described in related work, KGE models can mainly be divided into two categories: reconstruction-based and bilinear-based models. We explore the effectiveness of both two lines of works to extract features in our framework. Specifically, we leverage two classic KGE models as $\bm{f}$ to extract latent features in our experiment including TransE (reconstruction-based):
\begin{equation}
    \label{eq:kge_transe}
    \bm{f} = ||\hat{\bm{u}} + \bm{r} - \hat{\bm{v}}||_p,
\end{equation}
and ConvKB (bilinear-based):
\begin{equation}
    \label{eq:kge_convkb}
    \bm{f} = \bm{W}_1 \left( \text{ReLU}\big(\text{Conv}([\hat{\bm{u}};
    \bm{r}; \hat{\bm{v}}])\big) \right),
\end{equation}
where $\hat{\bm{u}}, \hat{\bm{v}}$ are the concatenation of top-$n$ component embeddings of node $u$ and node $v$ respectively; $||\cdot||_p$ denotes the $p$-norm operation; $[\cdot ;\cdot]$ denotes the concatenate operation;
$\text{Conv}(\cdot)$ indicates the convolutional layer with $M$ filters, and $\bm{W}_1 \in \mathbb{R}^{1 \times \frac{Mdn}{K}}$ is a trainable matrix.
In total, $\bm{f}$ is expected to give higher scores for valid triplets than invalid ones.

\vspace{-0.5em}
\paragraph{Network Embeddings (NEs)}
We first determine $\alpha^k_r$ according to the representations of node $u$ and node $v$ since NE usually does not provide explicit relations. Hence, $\alpha^k_r$ is calculated by performing a non-linearity transformation over the concatenation of $\bm{u}$ and $\bm{v}$:
\begin{equation}
\label{eq:attention}
    \alpha^{k}_r = \frac{\exp\left(\text{ReLU}(\bm{W}_2[\bm{u}^k;\bm{v}^k])\right)}{\sum_{j=1}^{K}\exp\left(\text{ReLU}(\bm{W}_2[\bm{u}^j;\bm{v}^j])\right)},
\end{equation}
where $\bm{W}_2 \in \mathbb{R}^{1 \times 2d}$ is a trainable matrix\footnote{Note that we also evaluate this definition on KG data, and the experimental results are presented in Appendix A.}.

Graph attention networks (GATs)~\cite{velivckovic2017graph} gather information from the node’s neighborhood and assign varying levels of importance to neighborhoods, which is a widely used and powerful way to learn embeddings for information networks.
Thus we leverage GATs as $\bm{f}$ to extract latent features for NE. Given a target node $u$ and its neighbors $\left\{v|v \in N_u\right\}$, we first determine the top-$n$ related components for each pair of nodes $(u, v)$ according to the attention weights $\alpha_r^k$. When updating the $k$-th component of $u$, a neighbor $v$ is considered if and only if the $k$-th component is in the the top-$n$ related components for the node pair $(u, v)$. 
In this way, we can thoroughly disentangle the neighbors of the target node into different components to play their roles separately. GATs are used to update each component as follows:
\begin{equation}
	\bm{u}^k =  \sum_{v \in N_u} \sigma (\bm{W}_3\bm{v}^k) \bm{W}_4^k \bm{v}^k ,
\end{equation}
where 
$\bm{W}_3 \in \mathbb{R}^{1 \times d}$ and $\bm{W}_4 \in \mathbb{R}^{h \times d}$ are two trainable matrices, $h$ is hidden size within GATs, and $\sigma$ is the softmax function which is used to calculate the neighbor's relative attention value in the $k$-th component.

\subsection{Updating Module}

\begin{figure}[!t]
	\centering
	\includegraphics[width=1.0\linewidth]{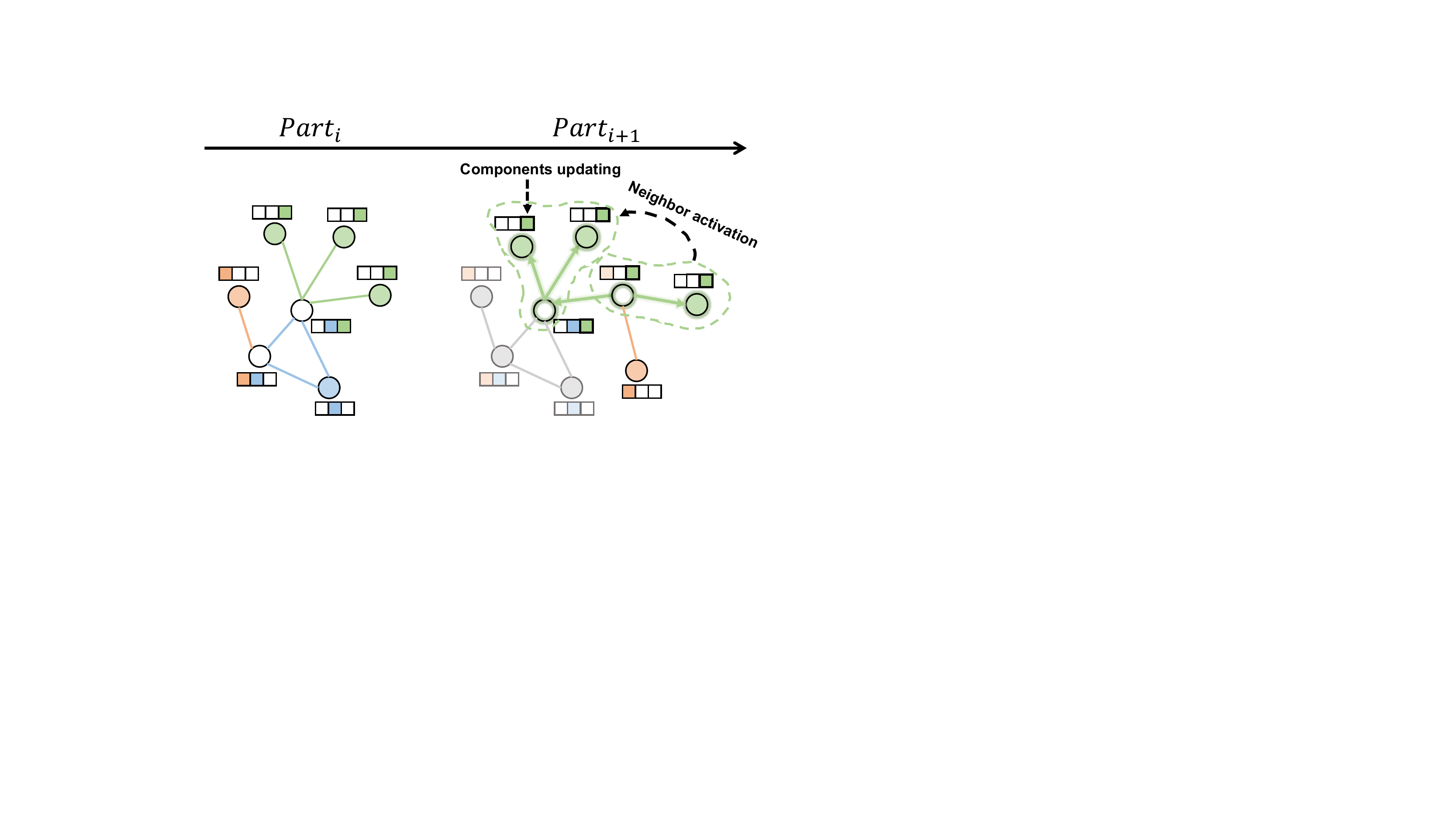}
	\caption{The updating module of \ourmodel, where different colors indicate disentangled components, and a node (white) may contain multiple components.}
	\label{fig:model_update}
	\vspace{-2mm}
\end{figure}

Now, the remaining problem is how to update the disentangled graph embedding when new relation triplets appear while preventing catastrophic forgetting. As shown in Figure~\ref{fig:model_update}, this process mainly includes two steps:

(1) \textbf{Neighbor activation}: \ourmodel needs to identify which relational triplets from $\mathcal{T}_1, ... , \mathcal{T}_{i-1}$ need to be updated. 
Since in the multi-relational data, nodes are not independent, and therefore a new relational triplet may have influence on the embeddings of nodes that not directly connect to it.
Inspired by that, for each relational triplet $(u,r,v)$, we activate both their direct and indirect neighbor triplets\footnote{1-order and 2-order neighbors are both considered in our experiments.}.
Specifically, neighbors of triplet $(u, r, v) \in \mathcal{T}_{i}$ refers to all triplets which contain node $u$ or node $v$ on the previous multi-relational graph $(\mathcal{T}_1, ... , \mathcal{T}_{i-1})$. 
In practice, adding all neighbors to $\mathcal{T}_i$ is computationally expensive occasionally since some nodes have very high degrees\footnote{The highest degree of FB15k-237 dataset is 7,614.}, i.e., they have a huge amount of neighbors. Therefore, we leverage a selection mechanism inspired by human's ability to learn procedural knowledge introduced in Section~\ref{sec_intro} and only update a few related neighbors: for each $(u,r,v)$, we only activate the neighbors with related semantic components (i.e., they share at least one component in their top-$n$ semantic components).

(2) \textbf{Components updating}: It is not necessary to update all semantic components of activated neighbors.
For example, if a relational triplet $(u', r', t') \in (\mathcal{T}_1, \cdots, \mathcal{T}_{i-1})$ is activated by $(u, r, t) \in \mathcal{T}_i$, we only need to update {\em the common components}, i.e., top-$n$ components, since the semantics of other components does not change. We use existing GE embedding method to update their features, as explained in our disentangle module.
Generally speaking, in each epoch, we iteratively train new relational triplets and relevant semantic aspects of activated neighbor relational triplets.
Through this training process, our model can not only effectively prevent catastrophic forgetting, but also learn the embeddings of new data.

\subsection{Training Objective}
\label{objective}
As mentioned before, for newly arrived multi-relational data $\mathcal{T}_i$, we iteratively train our model on $\mathcal{T}_i$ and its activated neighbor relational triplets. We denote loss functions of these two parts as $\mathcal{L}_{new}$ and $\mathcal{L}_{old}$ respectively. For KGE, we utilize soft-margin loss to train \ourmodel on link prediction task. The loss function $\mathcal{L}_{new}$ can be defined as:
\begin{equation}
\small
\label{eq:loss_1}
    \mathcal{L}_{new} = - \sum_{{(u,r,v)}\in \mathcal{T}_i\cup\mathcal{T}_i^*}  \log\left(1+ \exp\left(y \cdot \bm{f}(u, r, v)\right)\right),
\end{equation}
where $\mathcal{T}_i^*$ represents a set of relational invalid triplets for $\mathcal{T}_i$; $y = 1$ if $(u,r,t) \in \mathcal{T}_i$, otherwise, $y= -1$. For NE, we leverage a standard cross entropy loss according to GATs and train our model on node classification task. $\mathcal{L}_{new}$ can be formulated as follows:
\begin{equation}
\small
\label{eq:loss_2}
    \mathcal{L}_{new} = - \sum_{u\in N(\mathcal{T}_i) }\frac{1}{|C|} \sum_{c=1}^{|C|} y(c) \cdot \log\left(\sigma(\bm{W_5} \bm{u})\right), 
\end{equation}
where $c$ is node's class and $y(c)=1$ if the node label is $c$, otherwise $y(c)=0$; $N(\mathcal{T}_i)$ is the node set of $\mathcal{T}_i$, $|C|$ indicates the number of class, and $\bm{W_5} \in \mathbb{R}^{|C| \times d}$ is a trainable matrix. For KGE and NE, $\mathcal{L}_{old}$ can be defined in the same way with $\mathcal{L}_{new}$ on the selected old relational triplets set.

Intuitively, \textit{the less components a relation focuses on, the better the disentanglement  is}. Therefore, we add a constraint loss terms $\mathcal{L}_{norm}$ for $\mathcal{T}_i$~\footnote{The attention weights for the activated neighbor relational triplets are computed using the last checkpoint of the model and not updated during training $\mathcal{T}_i$.} to encourage the sum of the attention weights of the top-$n$ selected components to reach $1$, i.e., \begin{equation}
\label{eq:loss_rel}
    \small
    \mathcal{L}_{norm} = \sum_{(u, r, v) \in \mathcal{T}_i} (1 - \sum_{k}^{n}{\alpha}^k_r),
\end{equation}
where $n$ indicates the number of selected components. 

The overall loss function $\mathcal{L}$ of our proposed model is defined as follows:
\begin{equation}
\label{eq:loss_all}
    \mathcal{L} = \mathcal{L}_{old} + \mathcal{L}_{new} + \beta \cdot \mathcal{L}_{norm},
\end{equation}
where $\beta$ is a hyper-parameter.


\section{Experiments}
In this section, we evaluate our model on two popular tasks: \textit{link prediction} for knowledge graph and \textit{node classification} for information network.

\subsection{Datasets}

We conduct experiments on several continual learning datasets adapted from existing graph embedding benchmarks:

\textbf{KGE datasets.} We considered two link prediction benchmark datasets, namely FB15K-237~\cite{toutanova2015observed} and WN18RR~\cite{dettmers2018convolutional}.
We randomly split each benchmark dataset into five parts to simulate the real world scenarios, with each part having the ratio of $0.8:0.05:0.05:0.05:0.05$ respectively~\footnote{The characteristics of multi-relational graphs in real world scenarios are: 1) large-scale 2) the new multi-relational data is coming every day and small in scale proportional to the original size of these graphs.}. We further divide each part into training set, validation set and query set.
The statistics of FB15k-237 and WN18RR datasets are presented in Appendix C.

\textbf{NE datasets.} We conduct our experiments on three real-world information networks for node classification task: Cora, CiteSeer and PubMed~\cite{sen2008collective}. The nodes, edges and labels in these three citation datasets represent articles, citations and research areas respectively, and their nodes are provided with rich features.
Like KGE datasets, we split each dataset into four parts and the partition ratio is $0.7:0.1:0.1:0.1$. We further split train/validation/query set for each part. The statistics of Cora, CiteSeer and PubMed are presented in Appendix C.

\subsection{Experimental Settings}
\label{sec_exp}
We use Adam~\cite{kingma2014adam} as the optimizer and fine-tune the hyper-parameters on the validation set for each task.
We perform a grid search for the hyper-parameters specified as follows: the number of components $K \in \{2, 4, 6, 8, 10\}$, the number of top components $n \in \{2, 4\}$, node embedding dimension $d \in \{100, 200\}$ (note that relation embedding dimension in KG is $l = \frac{d \times n}{K}$, and $d$ is fixed to feature length in information network), initial learning rate $lr \in \{1e-3, 1e-4\}$, and the weight  of regulation loss $L_{norm}$ $\beta \in \{0.1, 0.3\}$.
The optimal hyper-parameters on FB15k-237 dataset are: $K = 8$, $n = 4$, $d = 200$, $lr = 1e-4$, $\beta = 0.3$; and the optimal hyper-parameters on WN18RR dataset are: $K = 4$, $n = 2$, $d = 200$, $lr = 1e-4$, $\beta = 0.1$. For the NE datasets, the optimal hyper-parameters are: $K = 8$, $n = 4$, $lr = 1e-3$, $\beta = 0.1$.

For a fair comparison, we implement the baseline models (TransE, ConvKB, and GATs) by ourselves based on released codes respectively, and use the same hyper-parameters as \ourmodel. For example, the embedding dimension for TransE and ConvKB are both $200$; the number of filters for ConvKB is $50$, and the number of heads in GATs is $8$.
We implement the baseline models of continual learning based on the toolkit~\footnote{\url{https://github.com/thunlp/ContinualRE}} released by  ~\citet{han2020continual}.
For fair comparison, we use the same memory size for both our model and baselines. For other hyper-parameters, we also follow the settings in~\citet{han2020continual}.

Following existing works~\cite{bordes2013translating}, we use the “Filtered” setting protocol, i.e., not taking any corrupted triples that appear in the KG into accounts, and the evaluation metrics of link prediction task include mean reciprocal rank (MRR) and proportion of valid test triplets in top-$10$ ranks (H@10). For node classification task, we use accuracy as our evaluation metric.
We use two settings to evaluate the overall performance of our \ourmodel after learning on all graphs: (1) \textbf{whole performance} calculates the evaluation metrics on the whole test set of all data; (2) \textbf{average performance} averages the evaluation metrics on all test sets. As average performance highlights the performance of handling catastrophic forgetting problem, thus it is the main metric to evaluate models.

\subsection{Baselines}
We compare our model with several baselines including two theoretical models to measure the lower and upper bounds of continual learning: 

(1) \textbf{Lower Bound}, which continually fine-tunes models on the new multi-relational dataset without memorizing any historical instances; 

(2) \textbf{Upper Bound}, which continually re-train models with all historical and new incoming instances. In fact, this model serves as the ideal upper bound for the performance of continual learning;
and several typical continual learning models:

(3) \textbf{EWC}~\cite{kirkpatrick2017overcoming}, which adopts elastic weight consolidation to add regularization on parameter changes. It uses Fisher information to measure the parameter importance, and slows down the update of those important parameters when learning new data;

(4) \textbf{EMR}~\cite{parisi2019continual}, a basic memory-based method, which memorizes a few historical instances and conducts memory replay.
Each time when new data comes in, EMR mixes memorized instances with new instances to fine-tune models; 

(5) \textbf{GEM}~\cite{lopez2017gradient}, which memorizes a few historical instances and adds a constraint on directions of new gradients to make sure that there is no conflict of optimization directions with gradients on old data.

\subsection{Overall Results}

\begin{table}[!t]
    \small
    \centering
    \begin{tabular}{ll|cc|cc}
    \toprule
         & Dataset & \multicolumn{2}{c|}{\textbf{FB15k-237} } & \multicolumn{2}{c}{\textbf{WN18RR} }  \\ 
         \midrule
         KGE & Model & W & A & W & A  \\  \midrule
         &Lower Bound  & 24.1 & 29.8 & 11.3 & 18.9 \\ 
         \cmidrule{2-6}
         &EWC  & 29.8 & 32.2 & 14.6 & 21.3 \\ 
         \multirow{2}{*}{TransE}&EMR  & 37.4 & 45.9 & 25.1 & 28.6 \\
         &GEM  & 38.3 & 46.1 & 25.3 & 28.9 \\ 
         \cmidrule{2-6}
         &\ourmodel  & \textbf{40.9} & \textbf{47.7} & \textbf{33.3} & \textbf{35.9} \\ 
         \cmidrule{2-6}
         &Upper Bound  & 43.5 & 49.7 & 50.8 & 49.5 \\
         \midrule
         &Lower Bound  & 13.2 & 19.3 & 6.3 &8.2 \\ 
         \cmidrule{2-6}
         &EWC  & 18.8 & 22.7 & 8.5 & 10.6 \\
         \multirow{2}{*}{ConvKB}&EMR  & 29.5 & 33.8 & 21.3 & 26.6 \\ 
         &GEM  & 28.0 & 35.0 & 25.3 & 27.4 \\ 
         \cmidrule{2-6}
         &\ourmodel  & \textbf{31.7} & \textbf{40.2} & \textbf{33.1} & \textbf{37.5} \\  
         \cmidrule{2-6}
         &Upper Bound  & 33.9 & 42.4 & 51.0 & 50.2 \\
         
    \bottomrule
    \end{tabular}
    \caption{H@10 (\%) results of models on two KG benchmarks. ``W'' stands for the \textbf{W}hole performance, and ``A'' stands for the \textbf{A}verage performance. }
    \label{tab:WN18RR, FB15k237}
\end{table}

\begin{table}[!t]
    \small
    \centering
    \begin{tabular}{l|cc|cc|cc}
    \toprule
         Dataset & \multicolumn{2}{c|}{\textbf{Cora} } & \multicolumn{2}{c|}{\textbf{CiteSeer} } &
         \multicolumn{2}{c}{\textbf{PubMed} }\\ 
         \midrule
         Model & W & A & W & A & W & A  \\  \midrule
         Lower & 61.2 & 60.5 & 60.9 & 61.8 & 82.3 & 81.8 \\
         \midrule
         EWC & 63.4 & 61.2 & 62.3 & 62.1 & 82.5 & 82.4 \\
         EMR & 72.4 & 73.9 & 66.8 & 68.9 & 83.1 & 83.0 \\  
         GEM & 75.3 & 76.1 & 70.9 & 70.2 & \textbf{85.5} & 84.2 \\ \midrule
         \ourmodel &  \textbf{78.1} &  \textbf{79.6} &  \textbf{72.1} &  \textbf{71.5} &  85.1 &  \textbf{85.0} \\
         \midrule
          Upper & 84.1 & 85.5 & 70.9 & 73.4 & 85.9 & 86.1\\
    \bottomrule
    \end{tabular}
    \caption{Accuracy results of models on three information network benchmarks (\%). ``Lower'' and ``Upper'' are abbreviations of the ``Lower Bound'' and ```Upper Bound'' baselines. }
    \label{tab:network}
    \vspace{-5mm}
\end{table}

Table~\ref{tab:WN18RR, FB15k237} and Table~\ref{tab:network} show the overall performance on both KGE and NE benchmarks under two evaluation settings. 
From the tables, we can see that:

(1) Our proposed \ourmodel model significantly outperforms other baselines and achieves state-of-the-art performance almost in all settings and datasets. It verifies the effectiveness of our disentangled approach in continual learning, which decouples the node embeddings into multiple components with respect to the semantic aspects, and only updates the corresponding components of graph embedding for new relational triplets.

(2) There is still a gap between our model and the upper bound. It indicates although we have proposed an effective approach for continual graph representation learning, it still remains an open problem deserving further exploration. 

(3) Although \ourmodel outperforms other baselines in almost all settings in three information network benchmarks, the performance gain is not as high as it is on the KG datasets.
The reason is that these three citation benchmarks are provided with rich node features, which would reduce the impact of topology changes. As can be seen, even the weakest Lower Bound achieves relatively high results close to Upper Bound.

\begin{figure}[!t]
	\centering
	\subfigure[FB15k-237]{
        \begin{minipage}[t]{0.49\columnwidth}
        \includegraphics[width=1.0\linewidth]{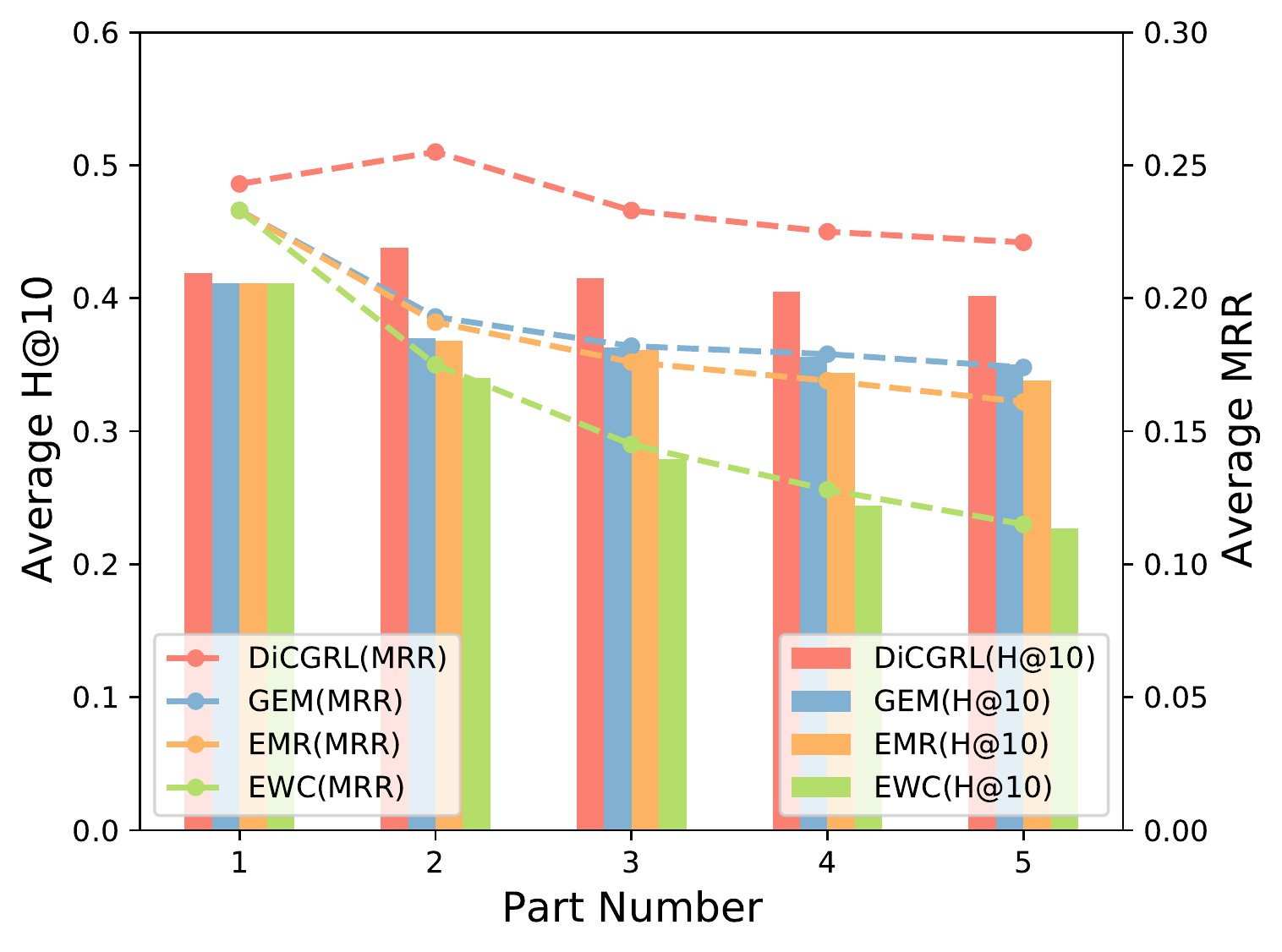}
        \vspace{-1mm}
        \label{fig:res_fb}
        \end{minipage}%
    }%
    \subfigure[WN18RR]{
        \begin{minipage}[t]{0.49\columnwidth}
        \includegraphics[width=1.0\linewidth]{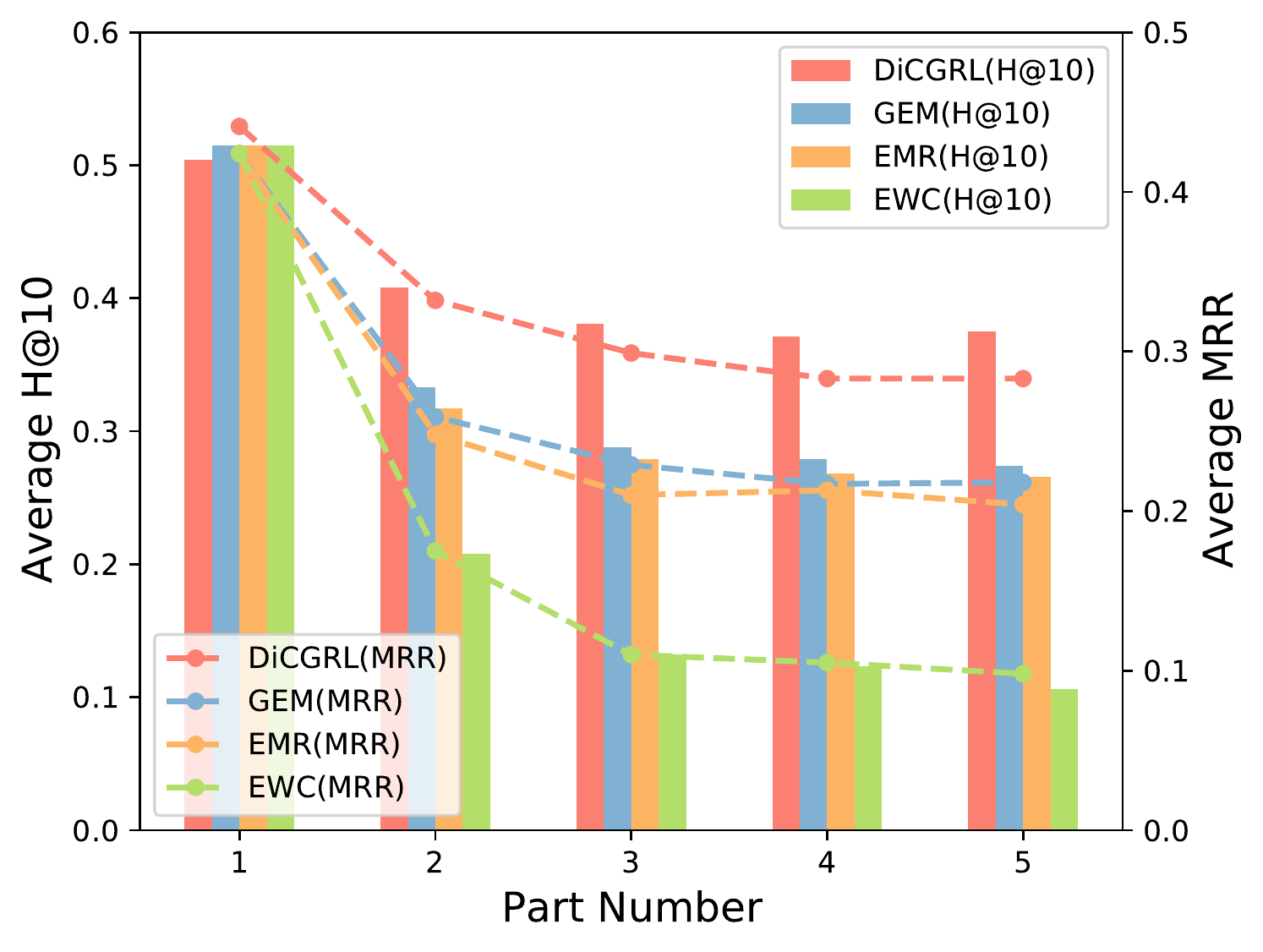}
        \vspace{-1mm}
        \label{fig:res_wn}
        \end{minipage}%
    }%
    \vspace{-2mm}
	\caption{Changes in H@10 and MRR with increasing knowledge graph data through the continual learning process, and the feature extraction method used in \ourmodel is ConvKB.}
	\vspace{-2mm}
	\label{fig:res_fb_wn}
\end{figure} 

\begin{figure}[!t]
	\centering
	\subfigure[Cora]{
        \begin{minipage}[t]{0.49\linewidth}
        \includegraphics[width=1.0\linewidth]{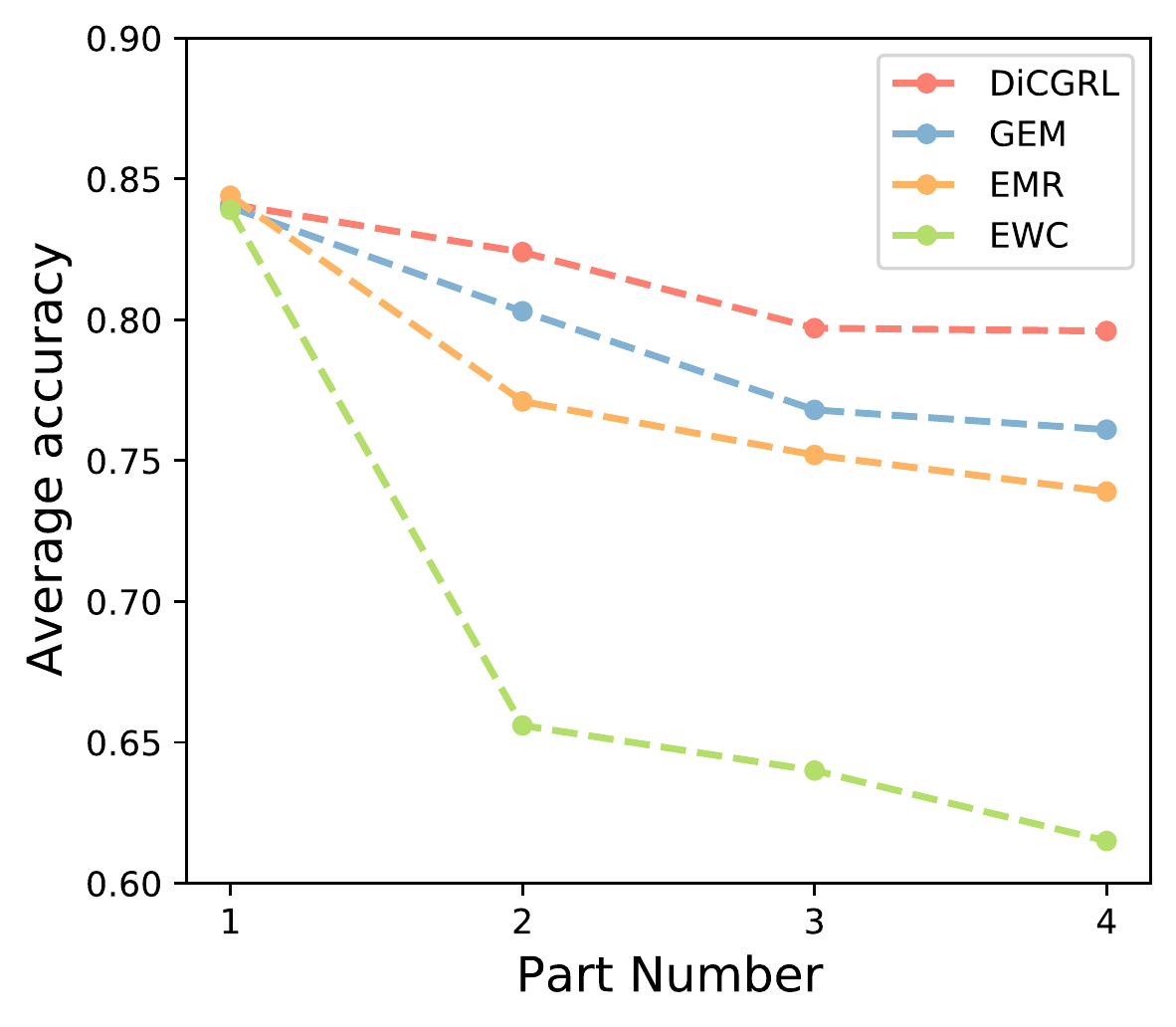}
        \vspace{-1mm}
        \label{fig:res_cora}
        \end{minipage}%
    }%
    \subfigure[CiteSeer]{
        \begin{minipage}[t]{0.49\linewidth}
        \includegraphics[width=1.0\linewidth]{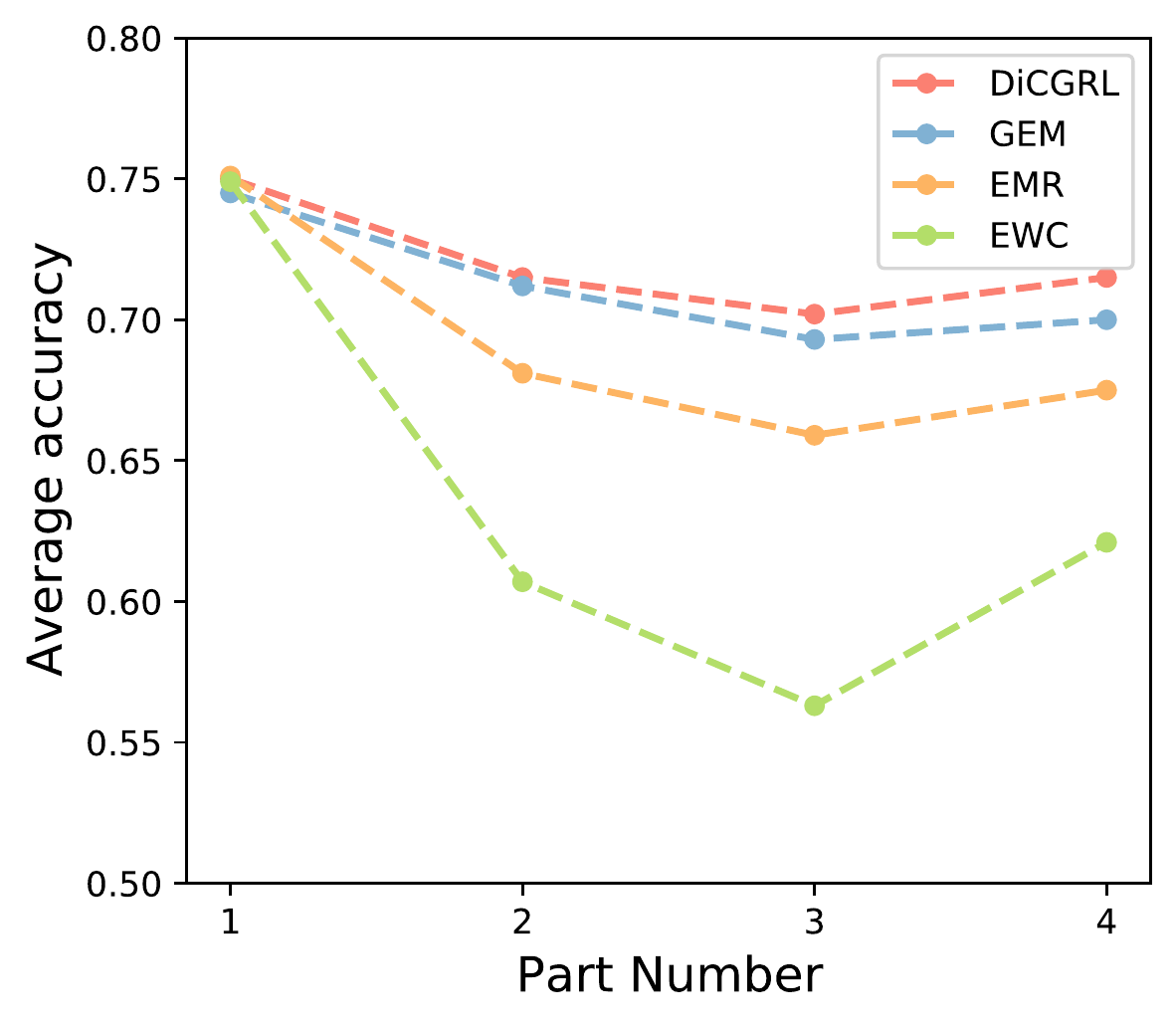}
        \vspace{-1mm}
        \label{fig:res_cite}
        \end{minipage}%
    }%
    \vspace{-1mm}
	\caption{Changes in accuracy with increasing information network data through the continual learning process. The result of PubMed dataset is presented in Appendix B.}
	\vspace{-2mm}
	\label{fig:res_net}
\end{figure} 

To further investigate how evaluation metrics change while learning new relational triplets, we show the average performance on the KG and NE datasets at each part in Figure~\ref{fig:res_fb_wn} and Figure~\ref{fig:res_net}. 
From the figures, we observe that: 

(1) With increasing numbers of new relational triplets, the performance of all the models in almost all the datasets decreases to some degree (CiteSeer may introduce some instability in random data splitting since this dataset is small). This indicates that catastrophically forgetting old data is inevitable, and it is indeed one of the major difficulties for continual graph representation learning. 

(2) The memory-based methods outperforms the consolidation-based methods, which demonstrates the memory-based methods may be more suitable for alleviating  catastrophic forgetting in multi-relational data to some extent. 

(3) Our proposed \ourmodel model achieves significantly better results compared to other baseline models. It indicates that disentangling relational triplets and updating dynamically selected components of relational triplets are more useful and reasonable than rote memorization of static examples from old multi-relational data.

\subsection{Hyper-Parameter Sensitivity}
In this section, we investigate the effect of the number of components $K$ and the top selected component number $n$, which are important hyper-parameters of our \ourmodel. These experiments are only performed on NE datasets, since the node embedding dimension $d$ can be affected by $K$ and $n$ in KG as introduced in Section~\ref{sec_exp}, which would make it difficult to make a fair comparison.

\begin{figure}[!t]
	\centering
	\subfigure[Effect of $K$]{
        \begin{minipage}[t]{0.49\linewidth}
        \includegraphics[width=1.0\linewidth]{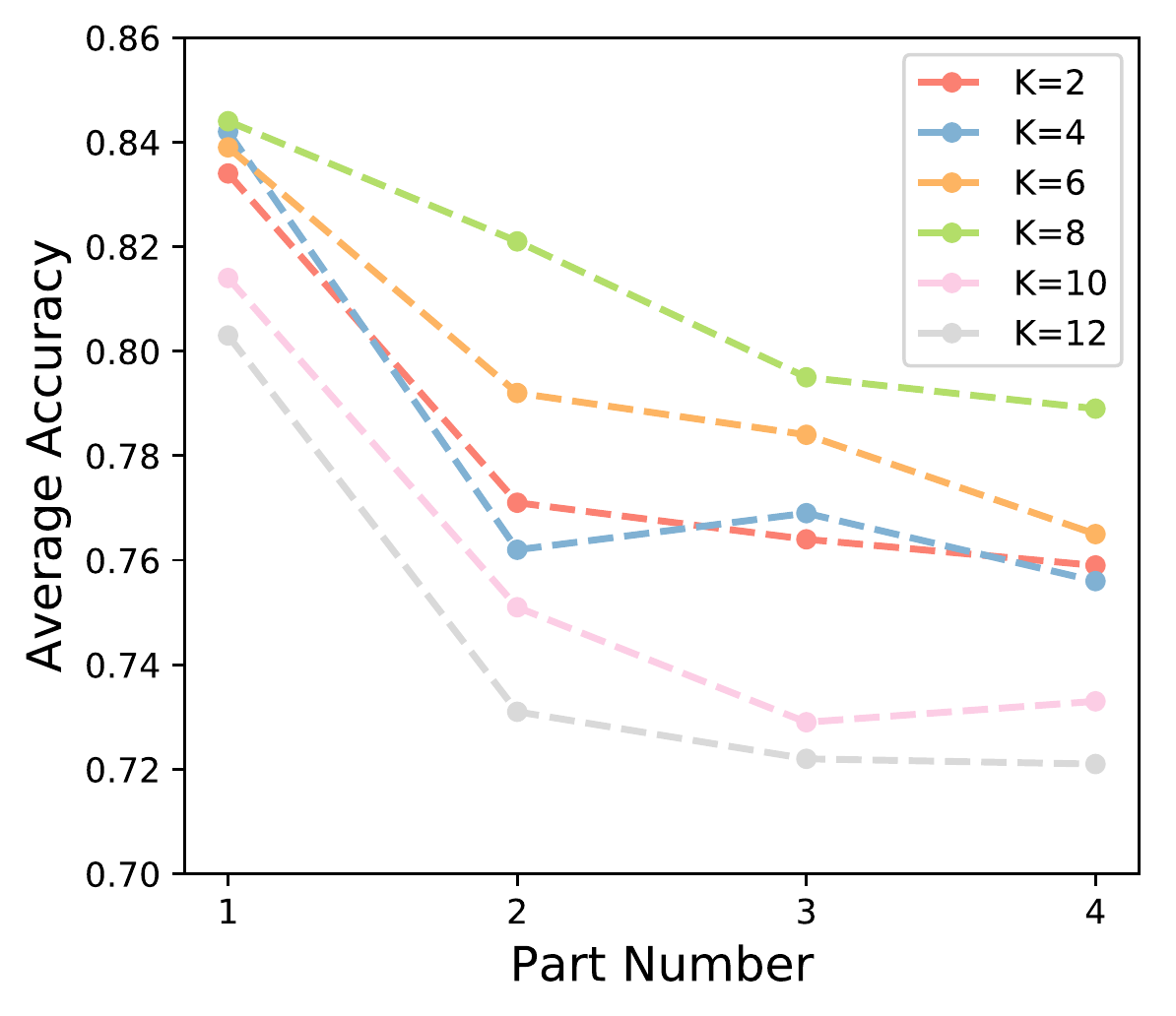}
        \vspace{-1mm}
        \label{fig:change_k}
        \end{minipage}%
    }%
    \subfigure[Effect of $n$]{
        \begin{minipage}[t]{0.49\linewidth}
        \includegraphics[width=1.0\linewidth]{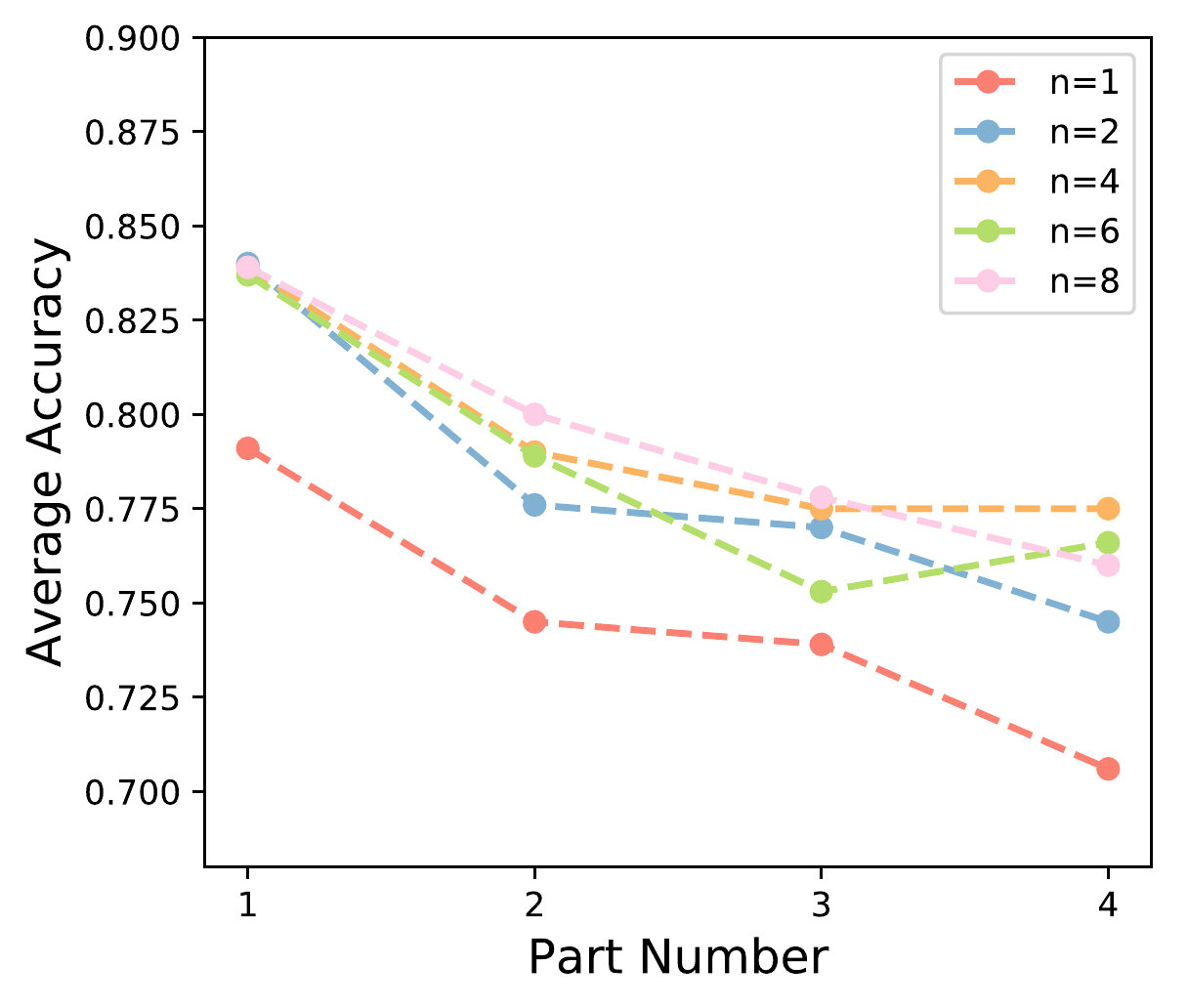}
        \vspace{-1mm}
        \label{fig:change_n}
        \end{minipage}%
    }%
	\caption{Hyper-parameter sensitivity of $K$ and $n$.}
	\label{fig:param}
\end{figure}

\textbf{Component Number $K$}: We use $n=2, \beta= 0.1$ to run \ourmodel on the Cora dataset with five different $K$ settings.
The results are illustrated in Figure~\ref{fig:change_k}.
From the figure, we find that overall the average accuracy raises when $K$ increases from $2$ to $8$, which suggests the importance of disentangling components.
However, when $K$ grows larger than $8$, the performance starts to decline. One possible reason is that the number of components is already larger than that of semantics aspects, making it harder to achieve a good disentanglement.
Therefore, we select the component number $K$ for each dataset on the development set, and for most dataset, we select $K=4$ or $K=8$.

\textbf{Top Selected Component Number $n$}: For a fair comparison, we set $K=8, \beta=0.1$ and vary $n$ on the Cora dataset. As shown in Figure~\ref{fig:change_n}, except for the case of $n = 1$, the other settings have comparable performance.
However, it can be seen that when $n = 4$, the average accuracy on the last task is the highest, which indicates that the model has the strongest ability to avoid catastrophic forgetting problem when $n=4$. 

\subsection{Efficiency Analysis}

\begin{figure}[!t]
	\centering
	\includegraphics[width=0.9\linewidth]{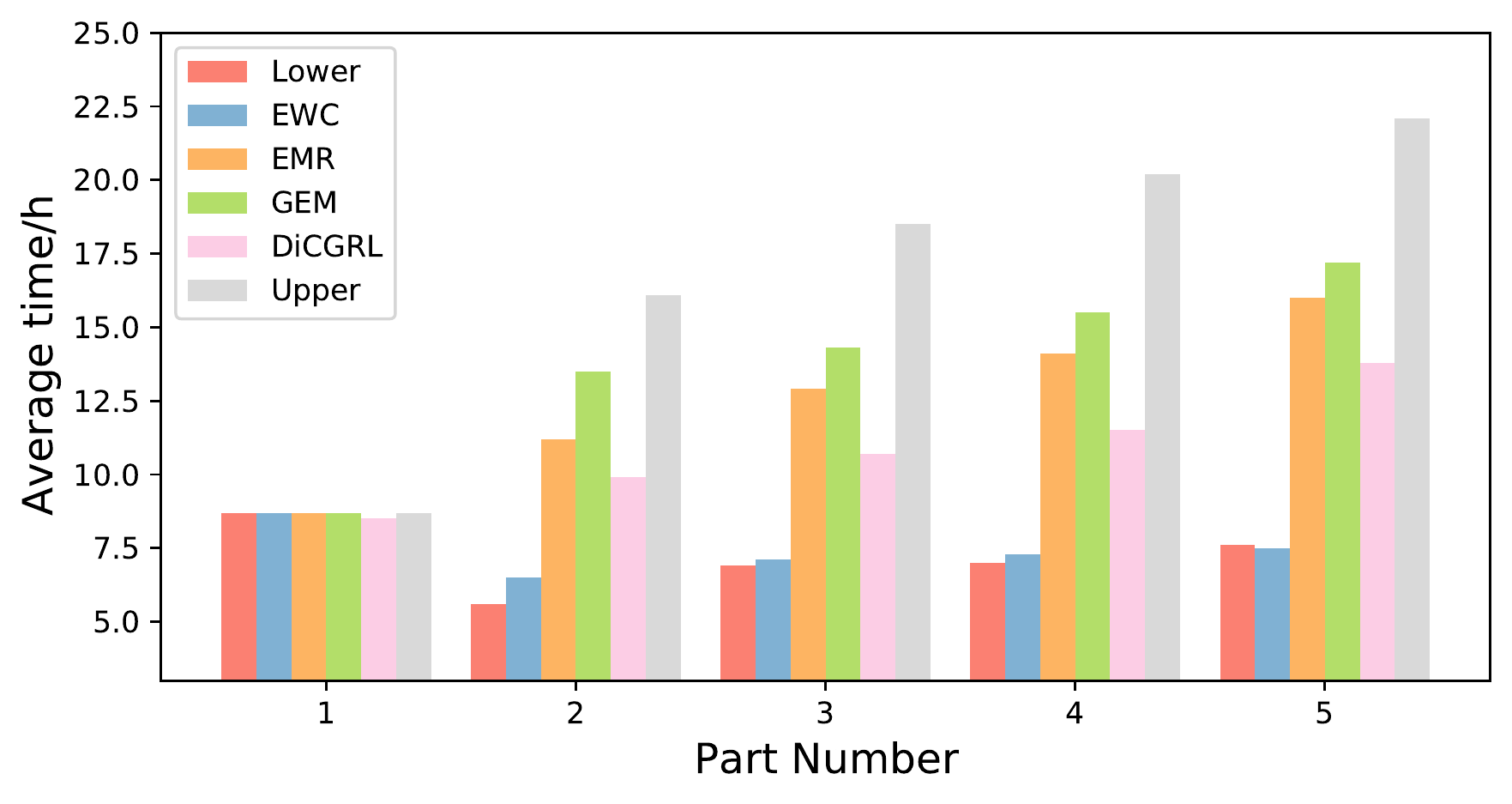}
	\vspace{-1mm}
	\caption{Training time (hour) on FB15k-237 datasets through the continual learning process.}
	\label{fig:cmp_time}
\end{figure}

We show the training time of different continual learning methods on the biggest benchmark FB15k-237, so as to highlight the efficiency gap in different methods. For a fair comparison, all algorithms use the same KGE method TransE. As shown in Figure~\ref{fig:cmp_time}, GEM takes much longer training time compared with other baselines, which is pretty close to Upper Bound.
Although our model also requires some previous data, it takes less time than GEM and EMR, which verifies the efficiency of our disentangled approach in continual learning.

\subsection{Case Study}

\begin{figure}[!t]
	\centering
	\subfigure[Example]{
        \begin{minipage}[t]{0.65\linewidth}
        \includegraphics[width=1.0\linewidth]{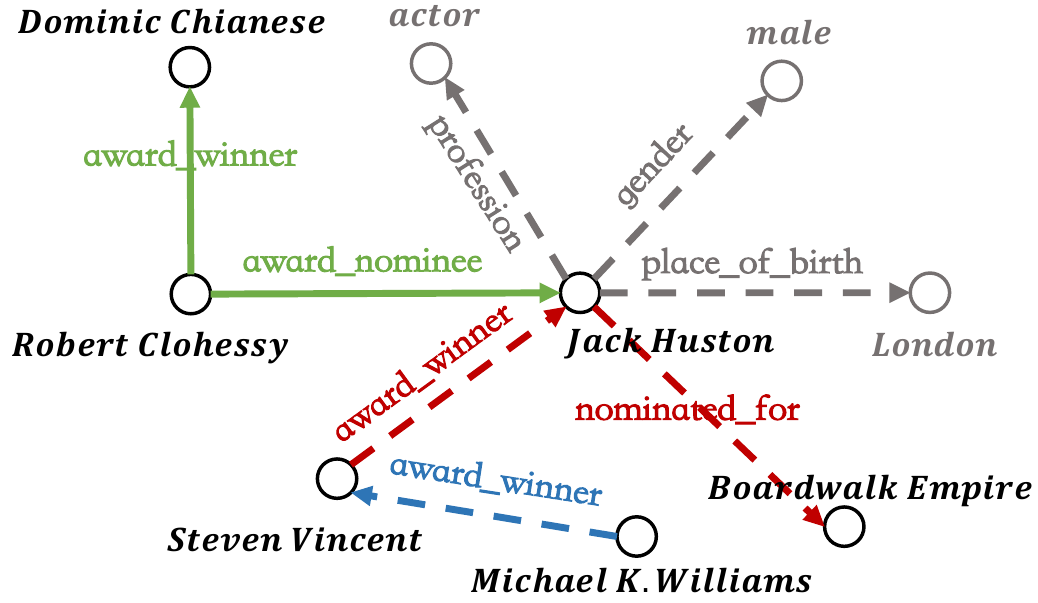}
        \vspace{-1mm}
        \label{fig:case_1}
        \end{minipage}%
    }%
    \subfigure[Visualization]{
        \begin{minipage}[t]{0.33\linewidth}
        \includegraphics[width=1.0\linewidth]{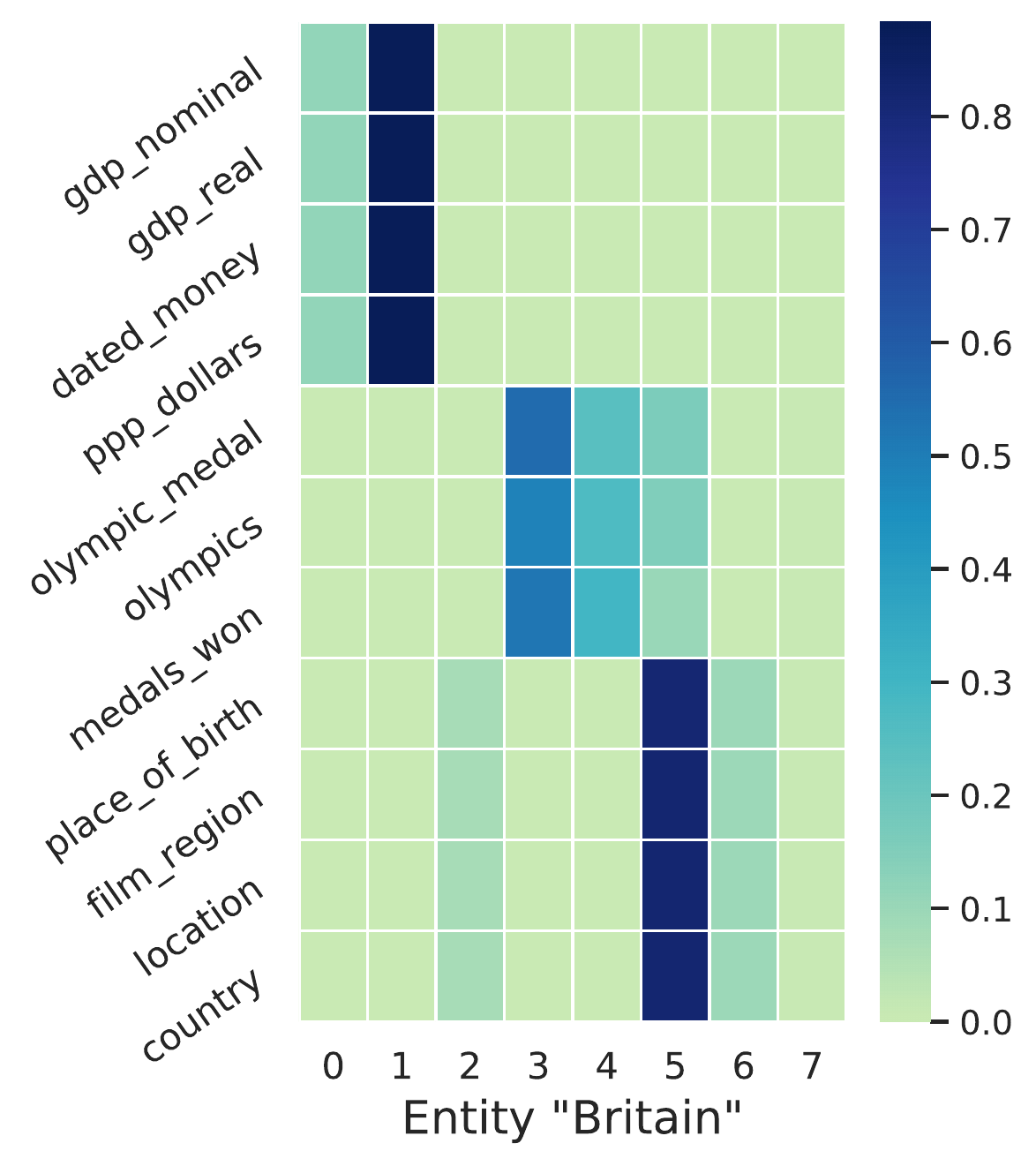}
        \vspace{-1mm}
        \label{fig:case_2}
        \end{minipage}%
    }%
    \vspace{-2mm}
	\caption{Figure(a) is an example of \ourmodel on the second part of FB15k-237 dataset, where solid line and dotted line represent newly arrived and previous relational triplets respectively. Red and blue color represents the activated 1-order and 2-order neighbors respectively. Figure(b) is the visualization of attention values on the entity {\em Britain}}
	\label{fig:case}
\end{figure}

In this section, we visualize an example case from FB15k-237 dataset (more readable than other datasets) to show that the activated neighbors in our updating module are in line with human commonsense. For example, as shown in Figure~\ref{fig:case_1} newly arrived relational triplets such as (\emph{Robert Clohessy}, \texttt{award\_nominee}, \emph{Jack Huston}) and (\emph{Robert Clohessy}, \texttt{award\_winner}, \emph{Dominic Chianese}), both related to ``award'' semantic aspects. Therefore, only ``award''-related neighbors of new triplets are updated, like (\emph{Jack Huston}, \texttt{nominated\_for}, \emph{Boardwalk Empire}). since \emph{Robert Clohessy} is also very likely to be related to the movie of \emph{Boardwalk Empire}. Meanwhile, relational triples of \texttt{place\_of\_birth} and \texttt{gender}, which are not related to ``award'', will not be updated.

Moreover, to verify the learned representation satisfies the intuition that different relations focus on different components of entities, we plot the attention values on the components of the entity {\em Britain} in Figure~\ref{fig:case_2}, where the y-coordinate is sampled relations that appear in the same triplets with ``{\em Britain}''.
We observe that semantically similar relations have similar attention value distributions. For example, relations ``\texttt{gdp\_nominal}'', ``\texttt{gdp\_real}'', ``\texttt{dated\_money}'', ``\texttt{ppp\_dollars}'', are all related to economics, relations ``\texttt{olympic\_medal}'', ``\texttt{olympics}'', ``\texttt{medal\_won}'' are all related to Olympics competitions.
These results demonstrate that the disentangled representations learned by our \ourmodel are semantically meaningful.

\section{Conclusion and Future Work}

In this paper, we propose to study the problem of continual graph representation learning, aiming to handle the streaming nature of the emerging multi-relational data. To this end, we propose a disentangled-based continual graph representation learning (\ourmodel) framework, inspired by human's ability to learn procedural knowledge. 
Extensive experiments on several typical KGE and NE datasets show that \ourmodel achieves consistent and significant improvement compared to existing continual learning models, which verifies the effectiveness of our model on alleviating the catastrophic forgetting problem. In the future, we will explore to extend the idea of disentanglement in the continual learning of other NLP tasks. 

\section*{Acknowledgments}
This work is supported by NSFC under Grant No. 61532001, National Key Research and Development Program of China under Grant No. 2018AAA0101902, and MOE-ChinaMobile Program under Grant No. MCM20170503.

\bibliography{emnlp2020}
\bibliographystyle{acl_natbib}

\appendix

\section{Results under Non-Continual Learning Settings}
Results under non-continual learning settings, i.e., using the entire training sets to train the models, are presented in Table~\ref{tab:whole_kg} and Table~\ref{tab:whole_net}. For a fair comparison, we also reproduce the results of baselines, and using the same optimal hyper-parameters as in Section Experimental Settings in the main paper. 
\ourmodel (T) and \ourmodel (C) indicate using ConvKB and TransE as feature extraction method for \ourmodel respectively. \ourmodel ($\cdot$)\_$\alpha$1 represents that $\alpha^k$ is calculated by performing a non-linearity transformation over the concatenation of $\bm{u}$ and $\bm{v}$ as doing in Network Embedding, and \ourmodel ($\cdot$)\_$\alpha$2 represents that $\alpha^k$ is calculated by performing a non-linearity transformation over the concatenation of $\bm{u}$, $\bm{r}$ and $\bm{v}$, 
since relations are an integral part of KGs.

\begin{table}[h!]
    \small
    \renewcommand
    \arraystretch{1.0}
    \centering
    \setlength{\tabcolsep}{7pt}
    \begin{tabular}{l|cc|cc}
    \toprule
         \multirow{2}{*}{\bf Model} & \multicolumn{2}{c|}{\textbf{WN18RR}} & \multicolumn{2}{c}{\textbf{FB15k-237} }  \\ 
         \cmidrule(lr){2-3}\cmidrule(l){4-5}
          & MRR & H@10 & MRR & H@10 \\ \midrule
         TransE & 22.6 & 50.1 & 29.4 & 46.5\\
         TransE$^*$ & 24.0 & \textbf{51.1} & 30.5 & 48.9 \\
         \ourmodel (T)\_$\alpha$1 & 17.3 & 40.8 & 29.0 & 46.8 \\
         \ourmodel (T)\_$\alpha$2 & 23.2 & 50.2 & 29.8 & 46.6 \\
         \ourmodel (T) & \textbf{24.2} & 50.4 & \textbf{30.9} & \textbf{49.6} \\ \midrule \midrule
         ConvKB & 24.9 & 52.4 & \textbf{24.3} & \textbf{42.1} \\
         ConvKB$^*$  & 40.4 & 53.6 & 22.5 & 41.3 \\
         \ourmodel (C)\_$\alpha$1 & 38.6 & 50.6 & 20.2 & 38.6 \\
         \ourmodel (C)\_$\alpha$2 & 39.2 & 49.4 & 20.7 & 38.2  \\
         \ourmodel (C) & \textbf{45.2} & \textbf{52.9} & 23.9 & 41.5 \\
    \bottomrule
    \end{tabular}
    \caption{Link prediction results on whole WN18RR and FB15k-237.
    The best score is in \textbf{bold}.
    Results of TransE and ConvKB are taken from \cite{nguyen2017novel} and \cite{sun2019re} respectively.
    $^*$ indicates reproduced by us.}
    \label{tab:whole_kg}
\end{table}

\begin{table}[h]
    \small
    \renewcommand
    \arraystretch{1.0}
    \centering
    \setlength{\tabcolsep}{8pt}
    \begin{tabular}{l|c|c|c}
    \toprule
          & \textbf{Cora} & \textbf{Citeseer} & \textbf{Pubmed} \\  \midrule
         GATs & 87.0 & 74.7 &  85.6  \\
         \ourmodel & \textbf{88.1} & \textbf{75.1} & \textbf{85.9} \\
    \bottomrule
    \end{tabular}
    \caption{Node classification results on three whole information networks.
    The best score is in \textbf{bold}.}
    \label{tab:whole_net}
\end{table}

From the tables, we can see that:

(1)~\ourmodel is comparable with our reproduced baselines, especially on the WN18RR with ConvKB as feature extraction method and FB15k-237 with TransE as feature extraction method, our \ourmodel performs better than vanilla GE methods. This phenomenon shows the effectiveness of our disentangled approach by decoupling the relational triplets in the graph into multiple independent components according to their semantic aspects.

(2)~As shown in Table~\ref{tab:whole_kg}, the performance of \ourmodel ($\cdot$)\_$\alpha$1 and \ourmodel ($\cdot$)\_$\alpha$2 are worse than \ourmodel, even worse than original baseline in some settings. This indicates that assigning global attention values for each relation as done in \ourmodel is an optimal option for KG datasets.

\section{PubMed Result}
\begin{figure}[h]
	\centering
	\includegraphics[width=0.6\linewidth]{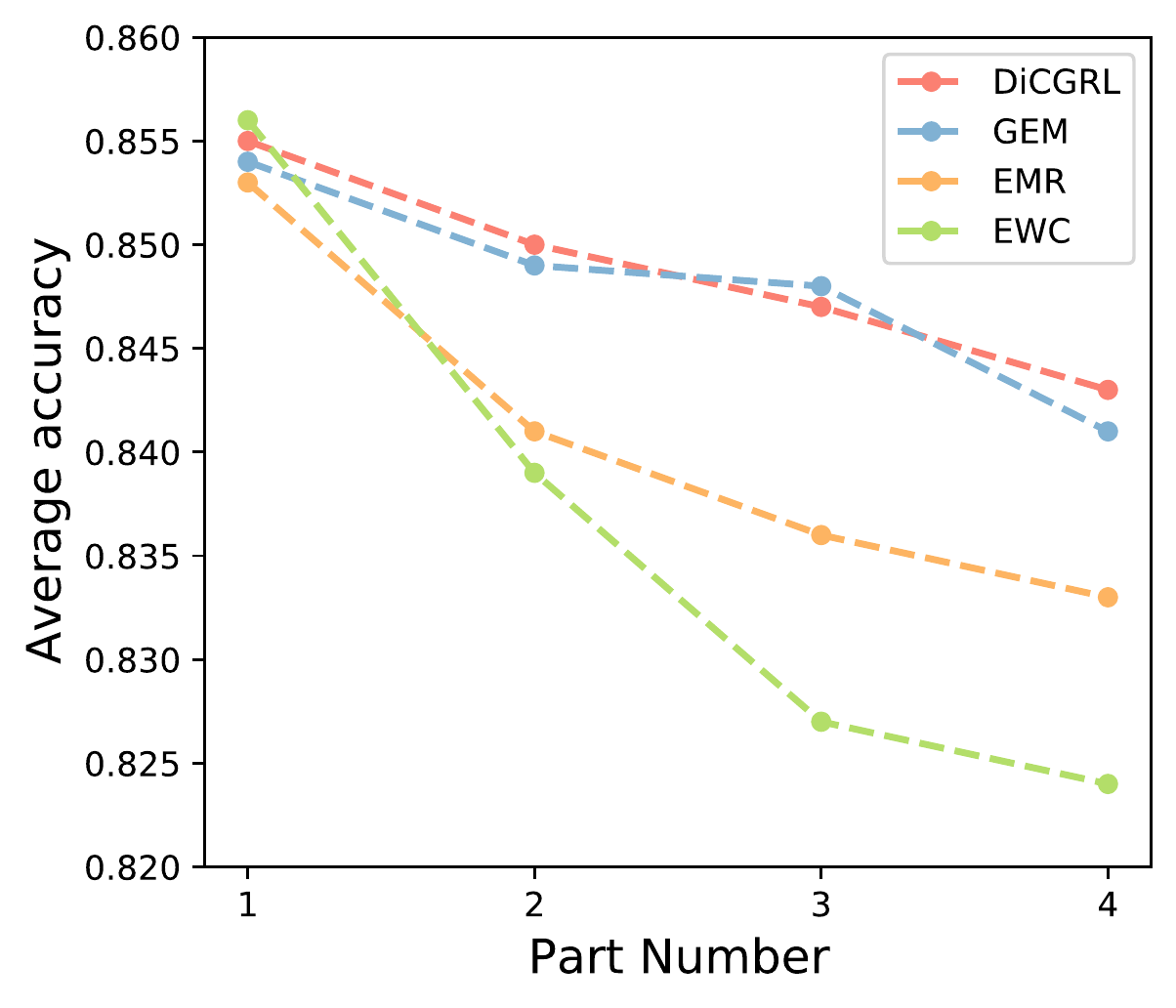}
	\caption{Changes in accuracy with increasing PubMed data through the continual learning process.}
	\label{fig:res_pubmed}
	\vspace{-5mm}
\end{figure}

The results of \ourmodel on the PubMed data is shown in Figure~\ref{fig:res_pubmed}.

\begin{table*}[t]
    \small
    \centering
    \setlength{\tabcolsep}{5pt}
    \begin{tabular}{l|rrrrr|rrrrr}
    \toprule
     \multirow{2}{*}{\bf Datasets} &\multicolumn{5}{c|}{\textbf{FB15k-237}} & \multicolumn{5}{c}{\textbf{WN18RR} }  \\ 
     \cmidrule(lr){2-6}\cmidrule(l){7-11}
     & P1 & P2 & P3 & P4 & P5 & P1 & P2 & P3 & P4 & P5  \\ \midrule
     \# Entities & 11,632 & 727 & 727 & 727 & 728 & 32,754 & 2047 & 2047 & 2047 & 2048 \\
     \# Relations & 236 & 236 & 236 & 236 & 237 & 11 & 11 & 11 & 11 & 11 \\ \midrule
     \# Train Set & 178,274 & 20,347 & 23,317 & 26,325 & 23,852 & 54,570 & 7,442 & 8,727 & 8,195 & 7,901 \\
     \# Validation Set & 11,726 & 1,263 & 1,382 & 1,635 & 1,529 & 1,899 & 269 & 272 & 309 & 285 \\
     \# Test Set & 13,681 & 1,556 & 1,642 & 1,801 & 1,786 & 1,970 & 273 & 321 & 284 & 286 \\ \midrule
     \# Accumulated Entities & 11,632 & 12,359 & 13,086 & 13,813 & 14,541 & 32,754 & 34,801 & 36,848 & 38,895 & 40,943 \\
     \# Accumulated Relations & 236 & 236 & 236 & 236 & 237 & 11 & 11 & 11 & 11 & 11 \\
    \bottomrule
    \end{tabular}
    \caption{Statistics of knowledge graph datasets. }
    \label{tab:Statistics_kg}
\end{table*}

\begin{table*}[t]
    \small
    \centering
    \setlength{\tabcolsep}{5pt}
    \begin{tabular}{l|rrrr|rrrr|rrrr}
    \toprule
     \multirow{2}{*}{\bf Datasets} &\multicolumn{4}{c|}{\textbf{Cora (\# Class = 7)}} & \multicolumn{4}{c}{\textbf{Citeseer (\# Class = 6)}} & \multicolumn{4}{c}{\textbf{PubMed (\# Class = 3)}}  \\ 
     \cmidrule(lr){2-5}\cmidrule(l){6-9}\cmidrule(l){10-13}
     & P1 & P2 & P3 & P4 & P1 & P2 & P3 & P4 & P1 & P2 & P3 & P4    \\ \midrule
     \# Nodes & 1,895 & 271 & 271 & 271 & 2,328 & 333 & 333 & 333 & 13,801 & 1,972 & 1,972 & 1,972  \\
     \# Edges & 2,475 & 764 & 1,046 & 1,144 & 2,347 & 685 & 742 & 958 & 22,287 & 6,272 & 7,533 & 8,246\\ \midrule
     \# Train Set & 568 & 81 & 81 & 81 & 698 & 99 & 99 & 99 & 4,140 & 591 & 591 & 591 \\
     \# Val Set & 379 & 54 & 54 & 54 & 466 & 67 & 67 & 67 & 2,760 & 395 & 395 & 395 \\
     \# Test Set & 948 & 136 & 136 & 136 & 1,164 & 167 & 167 & 167 & 6,901 & 986 & 986 & 986 \\ \midrule
     \# Acc\_Nodes & 1,895 & 2,166 & 2,437 & 2,708 & 2,328 & 2,661 & 2,994 & 3,327 & 13,801 & 15,773 & 17,745 & 19,717\\
     \# Acc\_Edges & 2,475 & 3,239 & 4,285 & 5,429 & 2,347 & 3,032 & 3,774 & 4,732 & 22,287 & 28,559 & 36,092 & 44,338 \\
    \bottomrule
    \end{tabular}
    \caption{Statistics of three citation datasets. }
    \label{tab:Statistics_net}
    \vspace{-3mm}
\end{table*}

\section{Dataset Statistics}
The statistics of FB15k-237 and WN18RR are presented in Table~\ref{tab:Statistics_kg}, where ``P$i$'' denotes the $i$-th part, ``\# Accumulated Entities'' and ``\# Accumulated Relations'' represent the cumulative entities and relations after each new part of multi-relational data is generated.
Statistics of Cora, Citeseer, and PubMed are presented in Table~\ref{tab:Statistics_net}, where ``\# Acc\_Nodes'' and ``\# Acc\_Edges'' represent the cumulative nodes and edges after each new part of multi-relational data is generated.

\end{document}